\documentclass[conference]{IEEEtran}
\IEEEoverridecommandlockouts
\usepackage{cite}
\usepackage{amsmath,amssymb,amsfonts}
\usepackage{algorithmic}
\usepackage{graphicx}
\usepackage{textcomp}
\usepackage{xcolor}
\usepackage{subcaption}
\usepackage[table]{xcolor}
\usepackage{booktabs}
\usepackage{enumitem}
\usepackage{subcaption}
\usepackage{balance}
\usepackage{multirow}
\usepackage{graphicx}
\usepackage{comment}
\def\BibTeX{{\rm B\kern-.05em{\sc i\kern-.025em b}\kern-.08em
    T\kern-.1667em\lower.7ex\hbox{E}\kern-.125emX}}
\begin{document}

\title{AGSR: Aperture-Guided Super-Resolution for Gaussian Splatting}

\author{\IEEEauthorblockN{Ayman Alkhateeb}
\IEEEauthorblockA{\textit{Dept. of Information Technology (ITEC)} \\
\textit{Alpen-Adria-Universität Klagenfurt}\\
Klagenfurt, Austria \\
Ayman.Alkhateeb@aau.at}
\and
\IEEEauthorblockN{Hadi Amirpour}
\IEEEauthorblockA{\textit{Dept. of Information Technology (ITEC)} \\
\textit{Alpen-Adria-Universität Klagenfurt}\\
Klagenfurt, Austria \\
Hadi.Amirpour@aau.at}
\and
\IEEEauthorblockN{Christian Timmerer}
\IEEEauthorblockA{\textit{Dept. of Information Technology (ITEC)} \\
\textit{Alpen-Adria-Universität Klagenfurt}\\
Klagenfurt, Austria \\
Christian.Timmerer@aau.at}
}

\maketitle

\begin{abstract}
3D Gaussian Splatting (3DGS) enables real-time novel view synthesis, but Gaussian rasterization attenuates fine-scale scene structure through the projected covariance of each primitive. We show that contracting these projected covariances at inference time can mitigate the effective bandwidth limitation induced by finite Gaussian footprints, motivating a multi-aperture rendering framework for super-resolution.

Based on this insight, we propose AGSR (Aperture-Guided Super-Resolution), a lightweight inference-time framework that progressively fuses multi-aperture renderings using uncertainty-guided feature aggregation and geometric conditioning from transmittance and depth cues. AGSR operates as a plug-in enhancement for pre-trained 3DGS models.

Experiments on the Mip-NeRF 360 benchmark show that AGSR outperforms lightweight image-space super-resolution baselines, with gains increasing as the underlying Gaussian representation deteriorates. The results provide evidence that covariance contraction exposes bandwidth-sensitive structural cues attenuated by native Gaussian rasterization. AGSR achieves these improvements while maintaining real-time performance with up to 139k parameters.
\end{abstract}


\section{Introduction}
\label{sec:introduction}

3D Gaussian Splatting (3DGS)~\cite{kerbl20233dgs} introduced an explicit scene representation based on anisotropic Gaussian primitives and efficient $\alpha$-compositing rasterization, enabling real-time novel view synthesis at competitive visual quality. Subsequent work has improved its robustness, scalability, and rendering quality~\cite{yu2023mipsplatting}. A fundamental limitation persists: the projected covariance of each Gaussian determines its screen-space footprint and therefore influences the local frequency response of the rasterized representation. Although the complete alpha-composited renderer is nonlinear, larger projected footprints attenuate high-frequency image content by increasing the spatial support of individual Gaussian kernels~\cite{yu2023mipsplatting,goodman2005fourier}. Consequently, high-frequency scene content, including sharp boundaries, fine surface texture, and sub-pixel geometric structure, can be attenuated by the finite spatial support of projected Gaussian primitives~\cite{goodman2005fourier,gonzalez2018digital}. Addressing this limitation, therefore, requires either increasing the representational capacity of the underlying scene model, often at additional computational and memory cost, or exploiting residual high-frequency cues that remain implicitly encoded within the rendered observations.

\begin{figure}[!t]
\centering
\includegraphics[width=0.90\columnwidth]{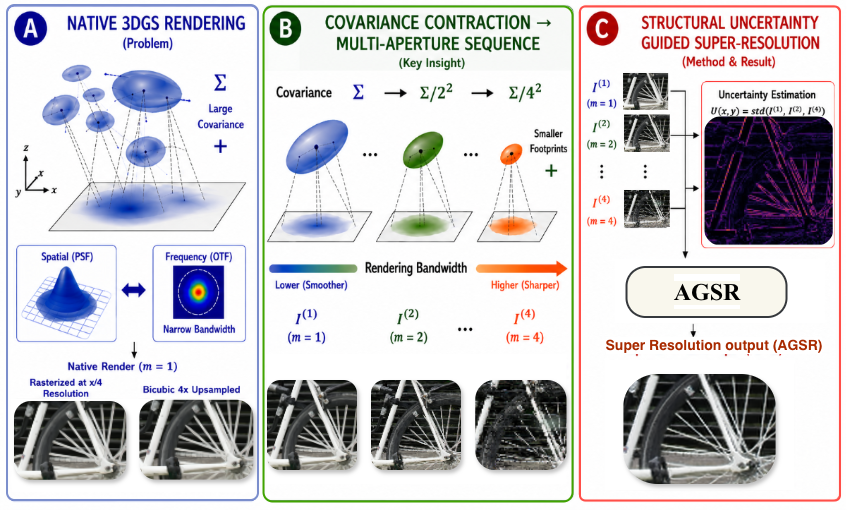}
\caption{
Projected Gaussian covariances influence the effective bandwidth of individual rasterization kernels. By systematically contracting the covariance, AGSR generates multi-aperture observations that reveal structural cues attenuated by native rasterization. The resulting uncertainty field guides geometry-aware super-resolution for enhanced detail reconstruction.}
\label{fig:theory_insight}
\end{figure}

The most direct response to this limitation is primitive densification~\cite{kerbl20233dgs}, which introduces additional Gaussian primitives to increase local representational capacity. While effective in principle, densification scales poorly: capturing increasingly fine geometric detail typically requires substantially more primitives, resulting in higher memory consumption, storage requirements, and rendering overhead~\cite{fan2024lightgaussian,papantonakis2024reduced3dgs,hanson2025pup3dgs}. Recent 3DGS super-resolution methods~\cite{yu2024gaussiansr,xie2024supergs,feng2024srgs,asthana2025splatsure} improve reconstruction fidelity through additional scene-level optimization and the incorporation of learned super-resolution priors. Although effective, these approaches increase reconstruction complexity, optimization time, and training overhead, limiting their practicality as lightweight inference-time solutions. Alternatively, lightweight 2D super-resolution networks~\cite{dong2016fsrcnn,shi2016espcn,sun2023safmn} can enhance rendered views with relatively modest inference overhead. However, because they operate solely in image space, they lack explicit geometric constraints and may produce perceptually plausible yet geometrically unconstrained high-frequency detail~\cite{blau2018perception,ren2025hallucination}.

We observe that the information required to recover high-frequency scene detail is not necessarily absent from the representation, but may instead be attenuated by the finite spatial extent of the rendering primitives, as shown in Fig.~\ref{fig:theory_insight}. Specifically, the projected covariance $\Sigma'$ of each Gaussian determines the screen-space footprint of the rasterized primitive~\cite{kerbl20233dgs,yu2023mipsplatting}. Although the 3DGS rasterization process is not strictly linear, larger projected footprints act analogously to low-pass smoothing operators, attenuating fine-scale image structure while preserving coarse scene content~\cite{goodman2005fourier,born1999principles,gonzalez2018digital}.

To probe this behavior, we systematically contract $\Sigma'$ during rasterization, generating a hierarchy of renderings with progressively smaller Gaussian footprints. We refer to this hierarchy as a \textit{multi-aperture rendering sequence}, where the term \textit{aperture} denotes the effective screen-space footprint scale of the rendering primitives, drawing inspiration from the role of aperture in controlling blur scale within optical imaging systems~\cite{goodman2005fourier}. The resulting image variations are not uniformly distributed across the scene: some regions remain comparatively stable across scales, whereas others exhibit pronounced sensitivity to footprint perturbations. We interpret this per-pixel variation as a structural uncertainty field, $\mathcal{U}(u,v)$, which localizes regions that are likely under-resolved by the native representation and can therefore benefit most from targeted enhancement.

The main contributions of this work are as follows:
\begin{itemize}

\item \textbf{A bandwidth-aware interpretation of Gaussian rasterization.}
We provide a bandwidth-oriented interpretation of Gaussian rasterization, showing how projected Gaussian covariances influence the effective frequency response of individual rasterization kernels and how covariance contraction can probe bandwidth-sensitive scene structure.

\item \textbf{A structural uncertainty field from multi-aperture rendering.}
We introduce a parameter-free uncertainty measure derived from cross-aperture rendering variation that localizes potentially under-resolved scene regions and provides a compact geometric prior for reconstruction.

\item \textbf{Aperture-guided super-resolution for Gaussian Splatting.}
We propose a lightweight architecture that progressively fuses multi-aperture observations conditioned on uncertainty, transmittance, and positional-encoded depth cues to enhance rendered 3DGS views without modifying the underlying representation.

\item \textbf{Validation across degradation and rendering scales.}
Experiments on Mip-NeRF 360 demonstrate increasing improvements under moderate and severe representation degradation.

\end{itemize}

\section{Theoretical Foundation}
\label{sec:theory}

Projected Gaussian footprints act as spatially varying reconstruction kernels whose frequency response influences the local frequency content of rendered images~\cite{yu2023mipsplatting}. For an isolated projected Gaussian,

\begin{equation}
G_i(p)
=
\exp
\left(
-\frac12
(p-\mu_i)^\top
\Sigma_i^{-1}
(p-\mu_i)
\right),
\end{equation}

where $\mu_i$ and $\Sigma_i$ denote the projected mean and projected covariance, respectively. The covariance controls the spatial extent of the footprint and, therefore, the local image resolution.

The Fourier transform of a two-dimensional Gaussian is

\begin{equation}
\widehat{G}_i(\omega)
=
2\pi
|\Sigma_i|^{1/2}
\exp
\left(
-\frac12
\omega^\top
\Sigma_i
\omega
\right).
\label{eq:gaussian_fourier}
\end{equation}

Since the Fourier transform of a Gaussian is also Gaussian, larger projected covariances produce broader spatial footprints and stronger attenuation of high-frequency components, whereas smaller covariances increase the effective bandwidth. Consequently, projected Gaussian footprints behave as local low-pass filters that attenuate high-frequency scene detail~\cite{gonzalez2018digital,pratt2007digital}. Practical 3DGS implementations further enforce a minimum projected footprint during rasterization~\cite{kerbl20233dgs}, increasing covariance and suppressing frequencies near the Nyquist limit.

\subsection{Aperture-Guided Covariance Contraction}

Motivated by this bandwidth limitation, we introduce covariance contraction factors
$m\in\{1,2,4\}$,
corresponding to progressively smaller rendering footprints.

The contraction is implemented by scaling the projected displacement

\begin{equation}
\Delta p^{(m)}
=
m(p-\mu_i),
\end{equation}

which is equivalent to replacing the projected covariance by

\begin{equation}
\Sigma_i^{(m)}
=
\frac{\Sigma_i}{m^2}.
\label{eq:sigma_contracted}
\end{equation}

Substituting Eq.\eqref{eq:sigma_contracted} into Eq.\eqref{eq:gaussian_fourier} yields

\begin{equation}
\widehat{G}_i^{(m)}(\omega)
=
2\pi
|\Sigma_i^{(m)}|^{1/2}
\exp
\left(
-\frac12
\omega^\top
\Sigma_i^{(m)}
\omega
\right).
\end{equation}

For $m>1$, since $\Sigma_i^{(m)} \prec \Sigma_i$, increasing $m$ may broaden the effective bandwidth and may reveal image structures attenuated at the native rendering scale. Although the renderings remain sampled on the same pixel lattice, the newly exposed responses appear as cross-aperture variations that provide evidence of potentially unresolved scene content.

Let $I^{(m)}(u,v)$ denote the rendering obtained at contraction level $m$. We quantify rendering sensitivity through the structural uncertainty field

\begin{equation}
\mathcal U(u,v)
=
\operatorname{Std}_{m\in\{1,2,4\}}
\left[
C^{(m)}(u,v)
\right].
\end{equation}

Large values of $\mathcal{U}(u,v)$ indicate regions whose appearance changes significantly with rendering bandwidth and are therefore likely to contain under-resolved scene structure. AGSR exploits this uncertainty field as a spatial prior for geometry-aware super-resolution.

\section{Proposed Method}

\begin{figure*}[!t]
\centering
\includegraphics[width=0.81\textwidth]{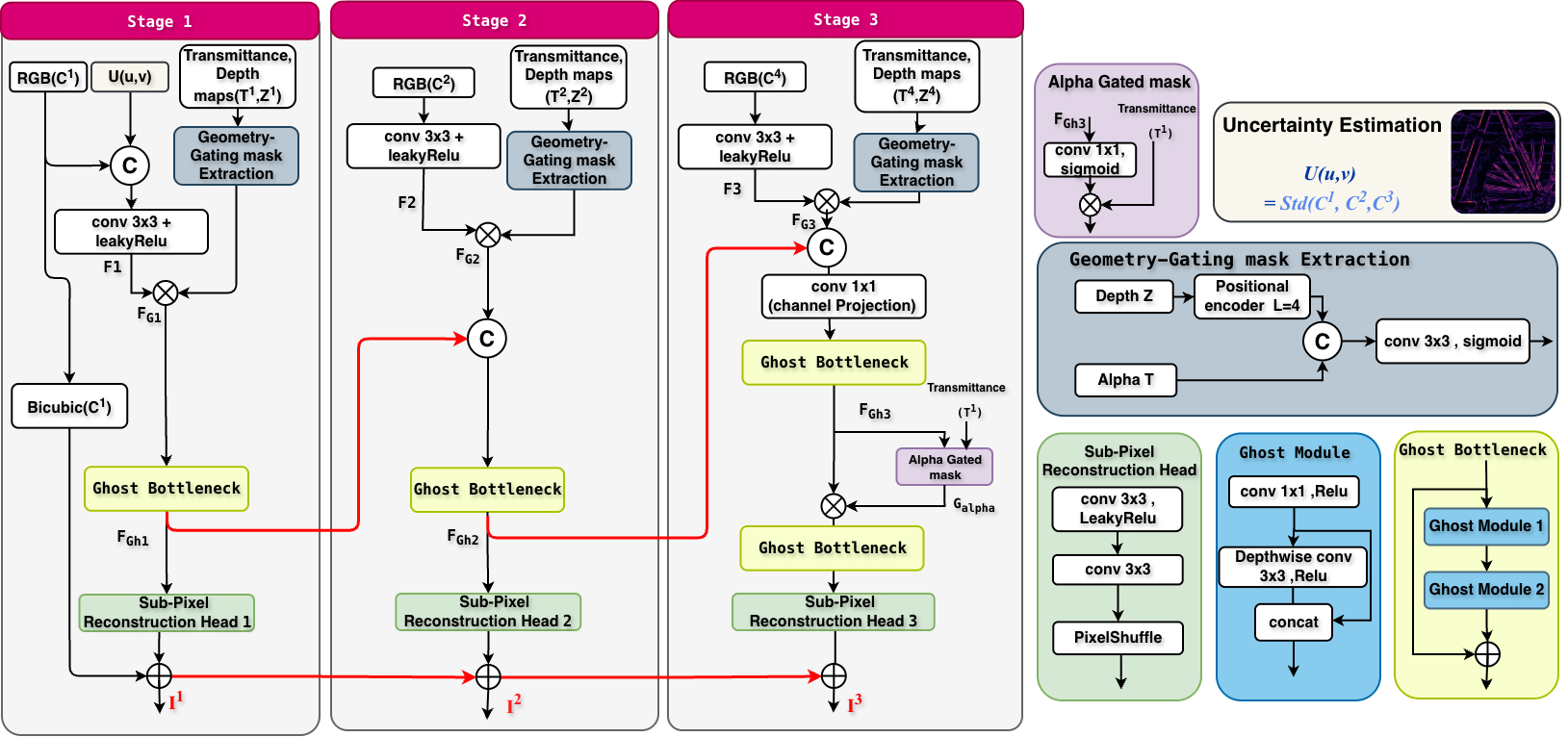}
\caption{
Overview of AGSR. Covariance contraction produces multi-aperture RGB, transmittance, and depth renderings with progressively smaller projected footprints and correspondingly higher effective bandwidth. Cross-aperture variations form a structural uncertainty field that guides a three-stage geometry-aware fusion network. Higher-bandwidth observations are fused sequentially and refined through residual PixelShuffle reconstruction to produce the final super-resolved output.
}
\label{fig:agsr_architecture}
\end{figure*}

The AGSR framework operates in three stages. First, a multi-aperture rendering sequence is generated by applying the covariance contraction hierarchy introduced in Section~\ref{sec:theory}, producing rendered observations $\{C^{(m)}, T^{(m)}, Z^{(m)}\}$ for $m\in\{1,2,4\}$ alongside the structural uncertainty field $\mathcal{U}(u,v)$. Second, these observations are processed by a geometry-aware progressive fusion network that reconstructs high-resolution novel views through successive aperture injection, geometric conditioning, and residual sub-pixel reconstruction (Figure~\ref{fig:agsr_architecture}).

\subsection{Multi-Aperture Rendering Inputs}

For each contraction level $m\in\{1,2,4\}$, the renderer produces an RGB image $C^{(m)}\in\mathbb{R}^{3\times H\times W}$, a rendered transmittance map $T^{(m)}\in\mathbb{R}^{1\times H\times W}$, and a rendered depth map $Z^{(m)}\in\mathbb{R}^{1\times H\times W}$. These signals are assembled into a multi-modal rendering tensor
\begin{equation}
\mathbf{X}^{(m)}
=
\bigl[
C^{(m)},
T^{(m)},
Z^{(m)}
\bigr]
\in
\mathbb{R}^{5\times H\times W}
\end{equation}
The native rendering at $m=1$ provides the base scene
representation, while contracted renderings at
$m=2$ and $m=4$ provide bandwidth-sensitive observations that may expose finer image structures attenuated at the native rendering scale.
\subsection{Structural Uncertainty Estimation}

The structural uncertainty field $\mathcal{U}(u,v)$ introduced in
Section~\ref{sec:theory} is computed over the RGB renderings
$\{C^{(m)}\}_{m\in\{1,2,4\}}$ of the multi-aperture sequence.
The resulting uncertainty map is concatenated with the base rendering
$C^{(1)}$ to form the Stage~1 network input
\begin{equation}
\bigl[
C^{(1)},
\mathcal{U}(u,v)
\bigr]
\in
\mathbb{R}^{4\times H\times W}.
\end{equation}

\subsection{Geometry-Gated Feature Modulation}

At each stage, the feature representation $F_k$ extracted from image $C^{(m_k)}$ in stage $k$, is modulated using geometric cues derived from the rendered transmittance map $T^{(m_k)}$ and the depth map $Z^{(m_k)}$. Following~\cite{mildenhall2020nerf,tancik2020fourier}, the depth map is transformed using a sinusoidal positional encoding
\begin{equation}
\gamma(Z^{(m)})
=
\left(
Z^{(m)},
\sin(2^l\pi Z^{(m)}),
\cos(2^l\pi Z^{(m)})
\right)_{l=0}^{L-1}.
\end{equation}

The resulting depth encoding comprises nine channels. Concatenating this with the transmittance map yields a 10-channel geometric-gating descriptor, which serves as a mask to modulate the feature hierarchy:

\begin{equation}
F_{G_k}
\leftarrow
F_k
\odot
\sigma
\Bigl(
\psi_k
\bigl(
[T^{(m_k)},\gamma(Z^{(m_k)})]
\bigr)
\Bigr).
\end{equation}

This operation suppresses feature responses in regions exhibiting low transmittance or weak geometric support while preserving responses near scene surfaces and depth discontinuities.

At the final stage, an additional Alpha Gate preserves features in reliable regions while suppressing features in unsupported regions.
\begin{equation}
\hat{\mathbf{F}}
=
\mathbf{F}_{Gh_3}
\odot
\mathbf{G}_{\alpha}
\qquad
\mathbf{G}_{\alpha}
=
\sigma\!\left(
\phi_\alpha(\mathbf F_{Gh_3})
\right)\odot \mathbf{T^1}
\end{equation}
Here $F_{Gh_3}$ indicates the stage $3$ GhostBottleneck output, while $\mathbf{T^1}$ is the transmittance map for $m=1$. The aim behind this refinement is to reduce the risk of hallucinated details in empty or low-confidence regions.

\subsection{Progressive Aperture Fusion Network}

Rather than concatenating all contraction levels at the
network input, AGSR fuses information from the
multi-aperture rendering sequence progressively across
three stages. Higher-bandwidth observations are injected
sequentially to refine an increasingly fine-scale image
structure.

The first stage operates on the base input $[C^{(1)},\mathcal{U}(u,v)]$, producing an initial feature representation $F_{Gh1}$. In subsequent stages, the RGB image rendered at contraction level $m_k\in\{2,4\}$ is encoded, geometry-gated, and concatenated with the accumulated feature hierarchy:
\begin{equation}
F_{Gh_k}
=\Phi_k 
\Bigl([F_{Gh_{k-1}},F_{G_k}]\Bigr),
\qquad k\in\{2,3\},
\end{equation}
where $m_2=2$ and $m_3=4$. Here, $\Phi_k(\cdot)$ denotes the fusion block at stage $k$, implemented using lightweight GhostBottleneck modules~\cite{han2020ghostnet}. Before the final fusion at stage $3$, a $1\times1$ convolution layer is employed for channel projection, reducing the feature dimensionality while preserving the most informative feature representations. This progressive fusion strategy enables the later stages to focus on increasingly localized detail refinement while maintaining computational efficiency.

\subsection{Residual Sub-Pixel Reconstruction}
Instead of independent reconstructions, AGSR concatenates deep features from the previous and current stages into an ESPCN-inspired sub-pixel head~\cite{shi2016espcn}. An upsampled spatial baseline $B$ is provided via bicubic interpolation, where $B = \operatorname{Bicubic}(C^{(1)})$ for $k=1$, and $B = I^{k-1}$ for $k > 1$. Each stage $k \in \{1,2,3\}$ then formulates the high-resolution output via a residual connection:
\begin{equation}
    \hat{I}^{(k)} = B + \mathrm{Up_{head}}(F_{Gh_{k}}),
\end{equation}
where $F_{Gh_{k}}$ denotes the features extracted by the Ghost Bottleneck module at stage $k$, which are subsequently processed by the upscale head ($\mathrm{Up_{head}}$).

\section{Results and Discussion}
\subsection{Experimental Setup}
\label{sec:setup}

During training, all three outputs ${I}^{(1)}$, ${I}^{(2)}$, and ${I}^{(3)}$ are supervised independently, providing deep supervision across the aperture hierarchy and encouraging meaningful intermediate reconstructions at every stage. During inference, AGSR can produce predictions from any stage, with ${I}^{(1)}$, ${I}^{(2)}$, and ${I}^{(3)}$ corresponding to one-, two-, and three-aperture inputs, respectively.

The upscaling factor, denoted by $s \in \{2, 4\}$, is governed exclusively by the PixelShuffle reconstruction heads, which generate $s^2$ intermediate channels before spatial rearrangement. Because the preceding feature extraction and fusion stages are strictly scale-agnostic, the AGSR framework flexibly accommodates multiple resolution targets without requiring architectural modifications.

To assess robustness under varying representation quality, we train separate 3DGS checkpoints using training images downsampled by factors $tf \in \{2,4,8\}$. Here, $tf$ controls the resolution of the training images, with larger values producing increasingly under-resolved Gaussian representations as a consequence of the lower-resolution training sets. We then evaluate two super-resolution scales, $s \in \{2, 4\}$. In both cases, AGSR renders and upsamples the output to match a fixed evaluation target.
All evaluation targets are obtained by downsampling the original Mip-NeRF 360 test images by a factor of $df=2$. AGSR inputs are then rendered at an additional factor of $s\in\{2,4\}$ below this fixed target resolution.
Separate AGSR model is trained for each combination of super-resolution scale $s$ and degradation factor $tf$, while sharing training data across all training scenes.
\begin{figure*}
\centering
\includegraphics[width=0.81\textwidth]{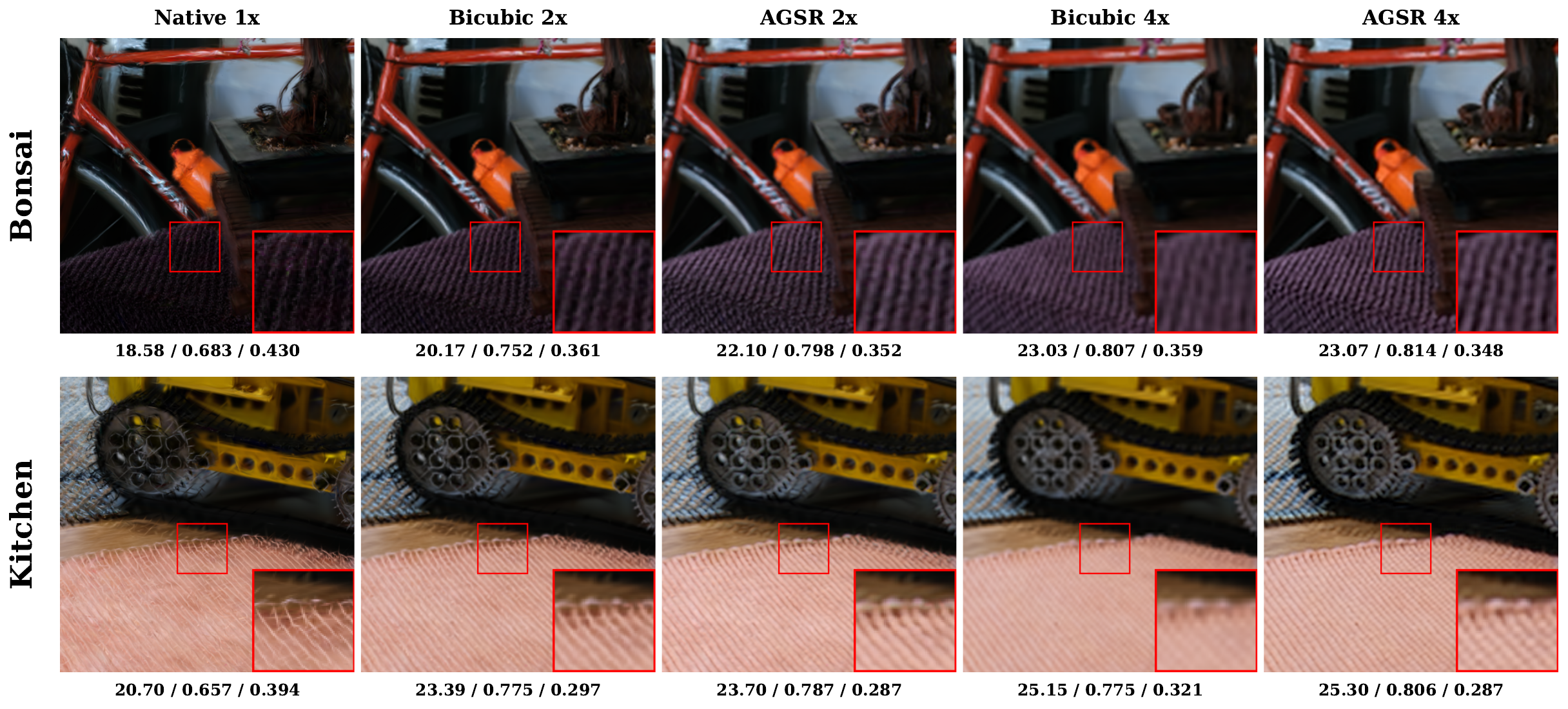}
\caption{\textbf{Visual comparison of $2\times$ and $4\times$ super-resolution on 3DGS representations trained on highly subsampled images ($tf=8$).} AGSR exploits structural cues revealed by multi-aperture rendering to recover fine-scale detail that is poorly reproduced by degraded Gaussian representations. Quantitative metrics (PSNR / SSIM / LPIPS) are reported below each crop.}
\label{fig:qualitative_tf8}
\end{figure*}

As a primary baseline, we include Native 3DGS~\cite{kerbl20233dgs} rendered directly at the target resolution, alongside a standard Bicubic upsampling baseline. This comparison is designed to evaluate whether low-resolution rendering followed by AGSR can outperform the direct high-resolution rasterization of the same frozen Gaussian representation.
Furthermore, we compare against two lightweight image-space super-resolution networks: SPAN~\cite{wan2024span} and SAFMN~\cite{sun2023safmn}. To ensure a fair, domain-adapted comparison, both networks are finetuned on rendered 3DGS patches from the training scenes using a learning rate of $1 \times 10^{-4}$ for 10 epochs. 

A separate baseline checkpoints are trained for each upscaling factor $s \in \{2,4\}$ and degradation level $tf \in \{2,4,8\}$, while the underlying 3DGS representations remain frozen throughout.

\begin{table}[!t]
\centering
\caption{Complexity analysis of the progressive aperture fusion network. We report the parameter count, computational complexity (GMACs), and full-pipeline inference speed (FPS) in both FP32 and FP16 precisions. Parameter counts and GMACs correspond to the cumulative deployed inference model required to produce each stage's output. FPS is evaluated targeting Full HD ($1920\times 1080$) and includes both 3DGS rasterization and network inference overhead.}
\label{tab:complexity}
\resizebox{\columnwidth}{!}{
\begin{tabular}{c|cccc|cccc}
\toprule
 & \multicolumn{4}{c|}{\textbf{AGSR $2\times$ ($960\times540 \rightarrow$ FHD)}} & \multicolumn{4}{c}{\textbf{AGSR $4\times$ ($480\times270 \rightarrow$ FHD)}} \\
\cmidrule(lr){2-5} \cmidrule(lr){6-9}
\textbf{Active Stages} & Params (K) & GMACs & FPS (FP32) & FPS (FP16) & Params (K) & GMACs & FPS (FP32) & FPS (FP16) \\
\midrule
1 & 21 & 10.6 & 40 & 43 & 36 & 4.7 & 103 & 112 \\
2 & 52 & 26.9 & 31 & 35 & 83 & 10.8 & 84 & 94 \\
3 & 92 & 47.6 & 22 & 27 & 139 & 18.0 & 64 & 76 \\
\bottomrule
\end{tabular}
}
\end{table}
\begin{table}[!t]
\centering
\caption{Quantitative comparison across multiple training degradation factors ($tf$). Best performance highlighted in \textbf{bold}.}
\label{tab:main_results}
\resizebox{\columnwidth}{!}{
\begin{tabular}{l|ccc|ccc|ccc}
\toprule
 & \multicolumn{3}{c|}{$tf=2$} & \multicolumn{3}{c|}{$tf=4$} & \multicolumn{3}{c}{$tf=8$} \\
\cmidrule(lr){2-4} \cmidrule(lr){5-7} \cmidrule(lr){8-10}
Method & PSNR$\uparrow$ & SSIM$\uparrow$ & LPIPS$\downarrow$ & PSNR$\uparrow$ & SSIM$\uparrow$ & LPIPS$\downarrow$ & PSNR$\uparrow$ & SSIM$\uparrow$ & LPIPS$\downarrow$ \\
\midrule
Native 3DGS & 28.74 & 0.857 & 0.260 & 25.83 & 0.799 & 0.299 & 21.15 & 0.631 & 0.437 \\
\midrule
\multicolumn{10}{c}{\textbf{Target Resolution: $2\times$ Super-Resolution}} \\
\midrule
Bicubic $2\times$ & 26.43 & 0.810 & 0.294 & 28.18 & 0.839 & 0.270 & 23.80 & 0.725 & 0.358 \\
SPAN-finetuned $2\times$ & 24.40 & 0.742 & 0.352 & 26.59 & 0.819 & 0.275 & 21.07 & 0.628 & 0.426 \\
SAFMN-finetuned $2\times$ & 27.16 & \textbf{0.830} & \textbf{0.281} & 26.62 & 0.820 & 0.268 & 22.27 & 0.674 & 0.387 \\
AGSR $2\times$ (Ours) & \textbf{27.20} & 0.827 & 0.287 & \textbf{28.28} & \textbf{0.845} & \textbf{0.264} & \textbf{25.49} & \textbf{0.764} & \textbf{0.349} \\
\midrule
\multicolumn{10}{c}{\textbf{Target Resolution: $4\times$ Super-Resolution}} \\
\midrule
Bicubic $4\times$ & 22.44 & 0.662 & 0.426 & 23.95 & 0.707 & 0.393 & \textbf{26.65} & 0.752 & 0.371 \\
SPAN-finetuned $4\times$ & 22.02 & 0.650 & 0.442 & 24.52 & 0.729 & 0.369 & 22.87 & 0.703 & 0.372 \\
SAFMN-finetuned $4\times$ & 23.55 & 0.692 & 0.395 & 24.32 & 0.720 & 0.364 & 23.24 & 0.705 & 0.363 \\
AGSR $4\times$ (Ours) & \textbf{24.62} & \textbf{0.724} & \textbf{0.397} & \textbf{25.41} & \textbf{0.752} & \textbf{0.368} & 26.63 & \textbf{0.766} & \textbf{0.360} \\
\bottomrule
\end{tabular}
}
\end{table}
\subsection{Training Objective}

The pre-trained 3DGS representation remains frozen, and only the AGSR network parameters are optimized. Training is performed for 40 epochs using Adam with cosine annealing warm restarts~\cite{loshchilov2017sgdr}. Multi-aperture images are rendered online, from which patch size $64\times64$ patches are extracted as input to the AGSR network and guided at all reconstruction stages via deep supervision.

The training objective combines Charbonnier reconstruction loss~\cite{lai2017deep}\cite{charbonnier1994two}, SSIM loss~\cite{wang2004ssim}, gradient consistency loss, Laplacian regularization, and a Fourier-domain loss:
\begin{equation}
\mathcal{L}
=
\lambda_c \mathcal{L}_{char}
+\lambda_s \mathcal{L}_{ssim}
+\lambda_g \mathcal{L}_{grad}
+\lambda_l \mathcal{L}_{lap}
+\lambda_f \mathcal{L}_{fft}.
\end{equation}
The Charbonnier term is weighted using a Sobel-derived edge map to emphasize structurally important regions, while the Fourier-domain term encourages recovery of high-frequency components attenuated by Gaussian rasterization. To facilitate stable optimization, AGSR employs a progressive curriculum schedule that gradually shifts supervision from early coarse reconstruction stages to the final high-frequency stage. Specifically, the weights of the first two stages decay from $1.0$ to $0.5$, while the final-stage weight increases from $0.2$ to $0.8$, encouraging a transition from global structure reconstruction to fine-scale detail recovery.

\begin{figure}[!t]
\centering
\includegraphics[width=1.0\columnwidth]{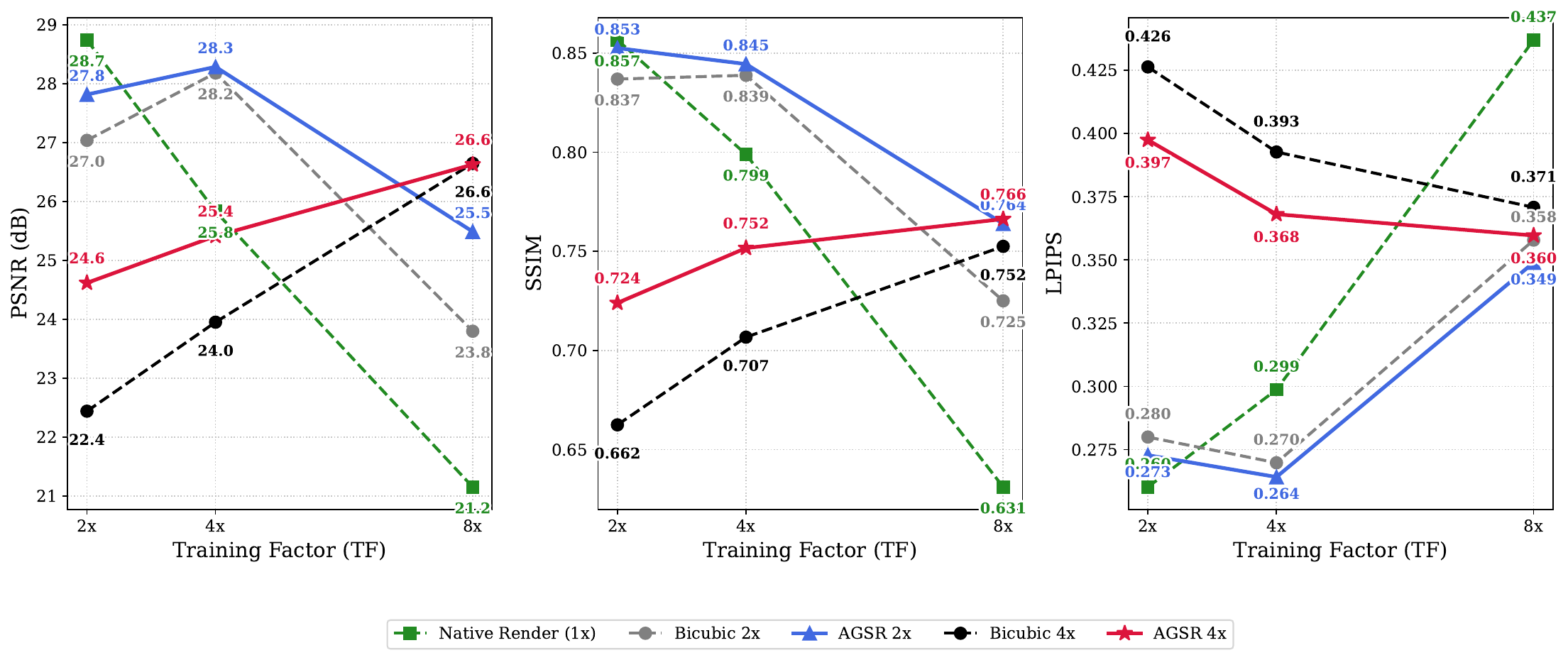}
\caption{Quantitative performance (PSNR, SSIM, and LPIPS) of AGSR compared with native 3DGS and bicubic baselines across varying training degradation factors ($tf$). As the training factor increases, the quality of the underlying Gaussian representation deteriorates substantially. AGSR becomes increasingly advantageous as the underlying representation degrades, with the strongest gains observed under moderate and severe degradation, outperforming the baselines across most settings and metrics.}
\label{fig:results}
\end{figure} 

\subsection{Complexity and Inference Speed}
Table~\ref{tab:complexity} demonstrates the high efficiency of AGSR across progressive fusion stages. All runtime measurements are obtained on an NVIDIA RTX A6000 using a batch size of 1. To ensure accurate timing, we report the average inference time over 300 frames following an initial warmup period. The network requires only 21K to 139K parameters, ensuring real-time or near real-time performance. Because the 4x task processes lower-resolution inputs (480x270), it significantly reduces spatial computation (GMACs) compared to the 2x task, resulting in fast inference speeds exceeding 60 FPS, even at Stage $3$. While adding active stages increases capacity, the computational cost remains acceptable across all configurations.

\subsection{Benchmark Results}
\label{sec:Benchmark Results}

Table~\ref{tab:main_results} summarizes performance across multiple training degradation factors. As the training set resolution decreases from 2 to 8, native 3DGS quality degrades substantially, with PSNR decreasing from 28.74 dB to 21.15 dB.

Qualitative results are shown in Fig.~\ref{fig:qualitative_tf8}. Under severe degradation, Bicubic upsampling can outperform direct high-resolution rasterization in PSNR and SSIM, as shown in Fig.~\ref{fig:results}. Unlike conventional 2D super-resolution, the low-resolution input is itself produced by a bandwidth-limited Gaussian rasterizer. Rendering at reduced resolution implicitly introduces stronger low-pass filtering, suppressing unstable high-frequency artifacts and aliasing generated by degraded Gaussian representations. An additional Bonsai scene ablation ($tf=8$) confirmed this effect: increasing the minimum projected footprint from 0.3 to 1.2 reduced the PSNR gap between native rendering and Bicubic upsampling from 1.75 dB to 0.22 dB.

At low degradation ($tf=2$) and moderate upscaling ($2\times$), SAFMN achieves marginally stronger SSIM (0.830 vs.\ 0.827) and LPIPS (0.281 vs.\ 0.287). This is consistent with the proposed theory: when the native Gaussian representation is already high-fidelity, covariance contraction exposes limited additional structure and $\mathcal{U}(u,v)$ provides weaker discriminative guidance. In this regime, the task approaches conventional image-space super-resolution, where specialized 2D SR networks remain highly competitive.

These results provide strong empirical support for the proposed framework. AGSR consistently achieves better overall performance under moderate and severe degradation while remaining competitive under mild degradation.

As the rendering becomes increasingly bandwidth-limited and under-resolved, AGSR demonstrates substantial improvements. At $2\times$ super-resolution with tf = 8, AGSR increases PSNR by 1.69 dB and SSIM by 0.039 compared with bicubic upsampling, and achieves a 4.34 dB PSNR gain over Native 3DGS. At $4\times$ super-resolution and $tf=8$, AGSR achieves superior perceptual quality with 0.766 SSIM and 0.360 LPIPS. The consistent gains obtained from multi-aperture observations indicate that covariance contraction exposes bandwidth-sensitive responses that are attenuated by native Gaussian rasterization. The ability of AGSR to exploit these cross-aperture variations suggests that they contain meaningful, scene-correlated structural cues rather than mere rendering artifacts.

\section{Conclusion}
We presented AGSR, a lightweight super-resolution framework for 3D Gaussian Splatting. By analyzing how projected Gaussian covariances influence the effective frequency response of the 3DGS rasterizer, we introduced covariance contraction as a mechanism for generating complementary bandwidth-sensitive observations from a frozen 3DGS representation. AGSR combines these multi-aperture observations with uncertainty, depth, and rendering-support cues through a lightweight progressive fusion network, without modifying or densifying the underlying Gaussian representation.

Experiments on Mip-NeRF 360 show that AGSR becomes increasingly advantageous as the underlying Gaussian representation becomes more under-resolved. With only 21K–139K trainable parameters, the framework provides consistent improvement under moderate and severe degradation while maintaining real-time or near-real-time inference depending on the operating configuration. Future work will investigate adaptive aperture selection, improved runtime efficiency, and stronger restoration under severely degraded scene representations.

\bstctlcite{IEEEexample:BSTcontrol}
\bibliographystyle{ieeetr}
\bibliography{refrences/ref}

\end{document}